\documentclass[runningheads]{llncs}
\usepackage{graphicx}
\usepackage{amsmath,amssymb} 
\usepackage{color}
\definecolor{blue1}{RGB}{102,178,255}
\definecolor{orange1}{RGB}{255,204,153}
\usepackage{dsfont}

\DeclareMathOperator*{\argmax}{argmax}
\newcommand{\1}[2]{\mathds{1}_{#1}(#2)}

\begin{document}
\pagestyle{headings}
\mainmatter

\def\ACCV20SubNumber{1018}  

\title{Discovering Multi-Label Actor-Action Association in a Weakly Supervised Setting} 
\titlerunning{Actor-Action Association}
%
\author{Sovan Biswas\inst{1}
\and
Juergen Gall\inst{1}
}
\authorrunning{Biswas, S. et al.}
%
\institute{University of Bonn, 53115 Bonn, Germany\\
\email{\{biswas,gall\}@iai.uni-bonn.de}}

\maketitle

\begin{abstract}
Since collecting and annotating data for spatio-temporal action detection is very expensive, there is a need to learn approaches with less supervision. Weakly supervised approaches do not require any bounding box annotations and can be trained only from labels that indicate whether an action occurs in a video clip. Current approaches, however, cannot handle the case when there are multiple persons in a video that perform multiple actions at the same time. In this work, we address this very challenging task for the first time. We propose a baseline based on multi-instance and multi-label learning. Furthermore, we propose a novel approach that uses sets of actions as representation instead of modeling individual action classes. Since computing the probabilities for the full power set becomes intractable as the number of action classes increases, we assign an action set to each detected person under the constraint that the assignment is consistent with the annotation of the video clip. We evaluate the proposed approach on the challenging AVA dataset where the proposed approach outperforms the MIML baseline and is competitive to fully supervised approaches.
\end{abstract}

\section{Introduction}

\begin{figure}[t]
    \begin{center}
       \includegraphics[trim=1.92cm 2cm 2cm 1.05cm,clip,width=0.32\linewidth]{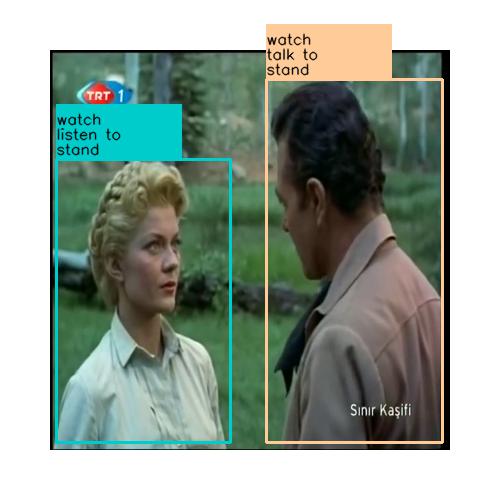}
    \end{center}
       \caption{The image shows a scene where two persons are talking. In this case there are two person that perform multiple actions at the same time. \textbf{Person A} indicated by the \textcolor{blue1}{\textbf{blue}} bounding box performs the actions \textit{Stand}, \textit{Listen to}, and \textit{Watch}. \textbf{Person B} indicated by the \textcolor{orange1}{\textbf{orange}} bounding box performs the actions \textit{Stand}, \textit{Talk to}, and \textit{Watch}. While in the supervised setting this information is also given for training, we study for the first time a weakly supervised setting where the video clip is only annotated by the actions     \textit{Stand}, \textit{Listen to}, \textit{Talk to}, and \textit{Watch} without any bounding boxes or associations to the present persons.}
    \label{fig:img1}
\end{figure}

In recent years, we have seen a major progress for spatially and temporally detecting actions in videos~\cite{gkioxari2015finding,hou2017tube,kalogeiton2017action,singh2017online,sun2018actor,Sun_2019_CVPR,biswas2019,girdhar2019video,wu2019long,feichtenhofer2019slowfast}. For this task, the bounding box of each person and their corresponding action labels need to be estimated for each frame as shown in Figure~\ref{fig:img1}. Such approaches, however, require the same type of dense annotations for training. Thus, collecting and annotating datasets for spatio-temporal action detection becomes very expensive. 

To alleviate this problem, weakly supervised approaches have been proposed~\cite{mettes2017localizing,soomro2017unsupervised,cheron2018flexible} where the bounding boxes are not given, but only the action that occurs in a video clip. Despite the promising results of the weakly supervised approaches for spatio-temporal action detection, current approaches are limited to video clips that predominantly contain a single actor performing a single action as in the datasets UCF 101~\cite{soomro2012ucf101} and JHMDB~\cite{jhuang2013towards}.
However, most real world videos are more complex and contain multiple actors performing multiple actions simultaneously. 
In this paper, we move a step forward and introduce the task of weakly supervised multi-label spatio-temporal action detection with multiple actors in a video.
The goal is to infer a list of multiple actions for each actor in a given video clip as in the fully supervised case~\cite{sun2018actor,Sun_2019_CVPR,biswas2019,girdhar2019video,wu2019long,feichtenhofer2019slowfast}. However, in the weakly supervised setting only actions occurring in each training video are known. Any spatio-temporal information about the persons performing these actions is not provided. This is illustrated in Figure~\ref{fig:img1} that shows two people standing and chatting. The video clip is only annotated by the four occurring actions \textit{Stand}, \textit{Listen to}, \textit{Talk to}, and \textit{Watch}. Additional information like bounding boxes or the number of present persons is not provided. In contrast to previous experimental settings for weakly supervised learning, the proposed task is much more challenging since a video clip can contain multiple persons, each person can perform multiple actions at the same time, and multiple persons can perform the same action. For instance, both persons in Figure~\ref{fig:img1} perform the actions \textit{Stand} and \textit{Watch} at the same time. 

In order to address multi-label spatio-temporal action detection in the proposed weakly supervised setup, we first introduce a baseline that uses multi-instance and multi-label (MIML) learning~\cite{zhou2006multi,zhou2012multi,yang2017miml}. Second, we introduce a novel approach that is better suited for the multi-label setting. Instead of modeling the class probabilities for each action class, we build the power set of all possible action combinations and model the probability for each subset of actions. Using a set representation has the advantage that we model directly the combination of multiple occurring actions instead of the probabilities of single actions. 
Since computing the probabilities for the full power set becomes intractable as the number of action classes increases, we assign an action set to each detected person under the constraint that the assignment is consistent with the annotation of the video clip. This is done by linear programming, which maximizes the overall gain across all plausible actors and action subset combinations. We evaluate the proposed approach on the challenging AVA 2.2 dataset~\cite{gu2018ava}, which is currently the only dataset that can be used for evaluating this task. In our experiments, we show that the proposed approach outperforms the MIML baseline by a large margin and that the proposed approach achieves $83\%$ of the mAP compared to a model trained with full supervision.        

In summary, the contribution of this paper is three-fold:
\begin{itemize}
    \item We introduce the novel task of weakly supervised multi-label spatio-temporal action detection with multiple actors.
    \item We introduce a first baseline for this task based on multi-instance and multi-label learning.
    \item We propose a novel approach based on an action set representation.
\end{itemize}
\section{Related Work}
\textbf{Spatio-Temporal Action Detection.} A popular approach for fully supervised spatio-temporal action detection comprises the joint detection and linking of bounding boxes~\cite{gkioxari2015finding,kalogeiton2017action,singh2017online,song2019tacnet}. These linked bounding boxes form tubelets which are subsequently classified. Recently, many methods~\cite{feichtenhofer2019slowfast,wu2019long,Feichtenhofer_2020_CVPR,Ji_2020_CVPR} use standard person detectors for actor localization and focus on learning implicitly or explicitly spatio-temporal interactions. All these approaches, however, require that each frame is annotated with person locations and corresponding action labels. Since such dense annotations are expensive to obtain on a large scale, recent approaches~\cite{gu2018ava,weinzaepfel2016towards,girdhar2019video} deal with temporally sparse annotations. Here, the action labels and locations are annotated only for a subset of frames. Even though there is a reduction in annotation, these methods still require person specific bounding boxes and their actions. Very few methods such as~\cite{mettes2017localizing,cheron2018flexible} explore the possibility of weakly supervised learning. Most of these methods such as~\cite{siva2011weakly,mettes2018spatio} use multiple instance learning to recognize distinct action characteristics. These works, however, consider the case where a single person performs not more than one action. 

\textbf{Actor-Action Associations.}
Actor-action associations have been key to identify actions both in a fully supervised and weakly supervised settings. \cite{Ghadiyaram_2019_CVPR} performs soft actor-action association using tags as pre-training on a very large dataset for fully supervised action recognition. With respect to weak supervision, a few approaches use movie subtitles ~\cite{bojanowski2013finding,laptev2008learning} or transcripts~\cite{kuehne2018hybrid,richard2018action} to temporally align actions to frames. In terms of actor-action associations for multiple persons, \cite{ramanathan2016detecting,Li_2020_WACV} associate a single action to various persons. To the best of our knowledge, our work is the first to perform multi-person and multi-label associations.

\textbf{Multi-Instance and Multi-Label Learning.} In the past, many MIML algorithms~\cite{nguyen2013multi,nguyen2010new} have been proposed. For example, \cite{zhou2012multi} propose the MIMLBoost and MIMLSVM algorithms based on boosting or SVMs. \cite{briggs2012rank} optimize a regularized rank-loss objective. MIML has been also used for different computer vision applications such as scene classification~\cite{zhou2006multi}, multi-object recognition~\cite{yang2017miml}, and image tagging~\cite{zha2008joint}. Recently, MIML based approaches have been used for action recognition~\cite{Li_2020_WACV,zhang2020multi}. 

\section{Multi-Label Action Detection and Recognition}
\label{sec:AlgoDetail}
Given a video clip with multiple actors where each actor can perform multiple actions at the same time as shown in Figure \ref{fig:img1}, the goal is to localize these actors and predict for each actor the corresponding actions. In contrast to fully supervised learning, where bounding boxes with multiple action labels are given for training, we address for the first time a weakly supervised setting where only a list of actions is provided for each video clip during training. This is a very challenging task as we do not know how many actors are present and each actor can perform multiple actions at the same time. This is in contrast to weakly supervised spatio-temporal action localization where it is assumed that only one person is in the video and that the person does not perform more than one action at a given point in time.   

In order to address this problem, we first discuss a baseline, which uses multi-instance and multi-label (MIML) learning \cite{zhou2006multi,zhou2012multi,yang2017miml}, in Section \ref{ssec:MIML}. In Section \ref{sec:method}, we will then propose a novel method which uses a set representation instead of a representation of individual actions. This means that we build from the annotation of a video clip the power set of all possible action combinations. For example, the power set $\Omega$ for the three action labels \textit{Listen}, \textit{Talk}, and \textit{Watch} is given by \{$\varnothing$, \{\textit{Listen}\}, \{\textit{Talk}\}, \{\textit{Watch}\}, \{\textit{Listen},\textit{Talk}\}, \{\textit{Listen},\textit{Watch}\},  \{\textit{Talk},\textit{Watch}\},
\{\textit{Listen},\textit{Talk},\textit{Watch}\}\}. We then assign one set $\omega_i \in \Omega \setminus \varnothing$ to each actor $a_i$ under the constraint that each action $c$ occurs at least once, i.e., $c \in \bigcup_i \omega_i$. Using a set representation has the advantage that we model directly the combination of multiple occurring actions instead of the probabilities of single actions.      

\section{Multi-instance and Multi-label (MIML) Learning}
\label{ssec:MIML}

One way to address the weakly supervised learning problem is to use multiple-instance learning. Since we have a multi-label problem, i.e., an actor can perform multiple actions at the same time, we use the concept of multi-instance and multi-label (MIML) learning \cite{zhou2006multi,zhou2012multi,yang2017miml}. 
We first use a person detector \cite{xie2017resnext} to spatially localize the actors in a frame $t$ and use a 3D-CNN such as I3D \cite{i3d} or Slowfast \cite{feichtenhofer2019slowfast} for predicting the action probabilities similar to fully supervised methods~\cite{girdhar2019video,wu2019long}. However, we use the MIML loss to train the networks.  

We denote by $A_t = \{a^t_1,a^t_2, \ldots, a^t_{n_t}\}$ the detected bounding boxes and by $f(a^t_i)$ the class probabilities that are predicted by the 3D-CNN. Let $Y$ be the vector which contains the annotations of the video clip, i.e., $Y(c)=1$ if the action class $c$ occurs in the video clip and $Y(c)=0$ otherwise. In other words, the bag $A_t$ is labeled by $Y(c)=1$ if at least one actor performs the action $c$ and by $Y(c)=0$ if none of the actors performs the action. The MIML loss is then given by
\begin{equation}
\mathcal{L}_{MIML} = \mathcal{L}\left(Y, \underset{i}{max}(f(a^t_{i})) \right) \label{eq:predopti}
\end{equation}
where $\mathcal{L}$ is the binary cross entropy. This means that the class probability should be close to one for at least one bounding box if the action is present and it should be close to zero for all bounding boxes if the action class is not present.   

\section{Actor-Action Association}
\label{sec:method}

\begin{figure}[t]
\centering
   \includegraphics[trim=0cm 0cm 0cm 0cm, clip,height=0.4\linewidth]{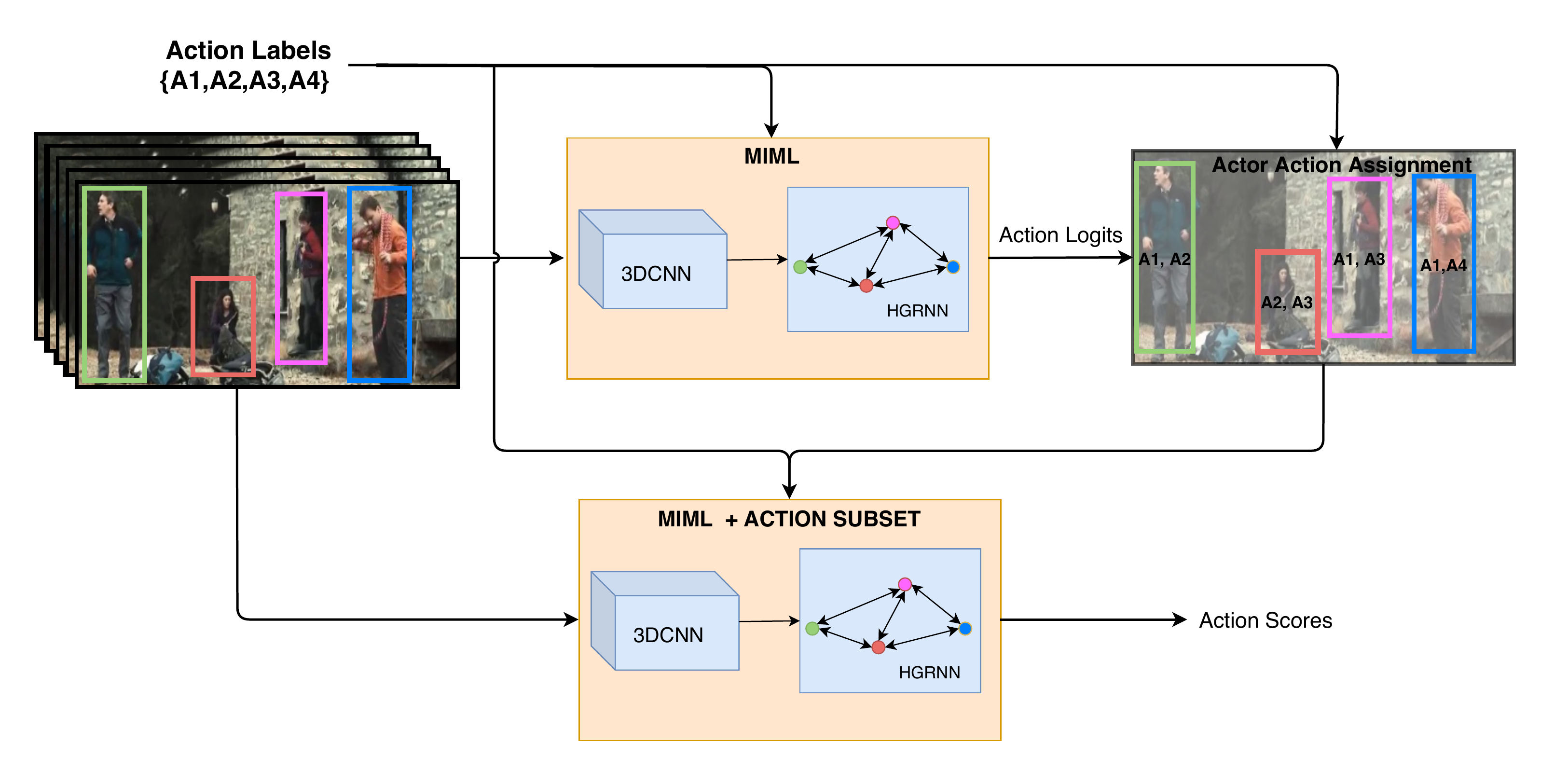}
   \caption{Overview of the proposed approach. Given a training video clip with action labels \{A1,A2,A3,A4\}, we first detect persons in the video. We then train a 3D CNN with a graph RNN that models the spatio-temporal relations between the detected persons using the MIML loss to obtain initial estimates of the action logits. During actor-action association, subsets of the action labels are assigned to each detected person. The training of the network is continued using the MIML loss and the actor-action associations.}
\label{fig:img3}
\end{figure}

While multi-instance and multi-label learning discussed in Section~\ref{ssec:MIML} already provides a good baseline for the new task of weakly supervised multi-label action detection, we propose in this section a novel method that outperforms the baseline by a large margin. As discussed in Section~\ref{sec:AlgoDetail}, the main idea is to change the representation from individual action labels to sets of actions. This means that we have one probability for a subset of actions $\omega \in \Omega$ instead of $C$ probabilities where $C$ is the number of action labels. We discuss how the probability of a set actions is estimated in Section~\ref{ssec:score}. Due to the weakly supervised setting not all combinations of subsets are possible for each video clip. We therefore assign an action set $\omega \in \Omega$ to each actor $a$ under the constraint that the assignment is consistent with the annotation of the video clip, i.e., each annotated action $c$ needs to occur at least once and actions that are not annotated should not occur. The assignment is discussed in Section~\ref{sec:asso}.  

Figure~\ref{fig:img3} illustrates the complete approach. As described in Section~\ref{ssec:MIML}, we use a 3D CNN such as I3D \cite{i3d} or Slowfast \cite{feichtenhofer2019slowfast}. Since the actors in a frame often interact with each other, we use a graph to model the relations between the actors. The graph connects all actors and we use a graph RNN to infer the action probabilities for each actor based on the spatial and temporal context. In our approach, we use the hierarchical Graph RNN (HGRNN) \cite{biswas2019} where the features per node are obtained by ROI pooling over the 3D CNN feature maps. The HGRNN and 3D CNN are learned using the MIML loss \eqref{eq:predopti}. From the action class probabilities, we infer the action set probabilities as described in Section~\ref{ssec:score} and we infer the action set for each actor as described in Section~\ref{sec:asso}. Finally, we train the HGRNN and the 3D CNN based on the assignments. This will be discussed in Section~\ref{sec:train}.

\subsection{Power Set of Actions}
\label{ssec:score}

In principle, we could modify our network to predict the probability for each subset of all action classes instead of the probabilities for all action classes. However, this is infeasible since the power set of all actions is very large. If $C$ is the number of actions in a dataset, the power set for all actions consists of $2^{C}$ subsets. Already with 50 action classes, we would need to predict the probabilities for over one quadrillion subsets.
Instead, we use an idea that was proposed for HEX graphs \cite{deng2014large} where the probabilities of a hierarchy are computed from the probabilities of the leave nodes. While we do not use a hierarchy, we can compute the probability of a subset of actions from the predictions of a network for individual actions. 

Let $s_{c} \in (-\infty,\infty)$ denote the logit that is predicted by the network for the action class $c$. The probability of a subset of actions $\omega$ can then be computed by
\begin{equation}
	p_{\omega} = \frac{\exp\left(\sum_{c \in \omega} s_c\right)}{\sum_{\omega'}\exp\left(\sum_{c \in \omega'} s_c\right)}. \label{eq:prob}
\end{equation}
The normalization term, however, is still infeasible to compute since we still need to sum over all possible subsets ($\omega'$) for the dataset.  

Since our goal is the assignment of a subset of actions $\omega$ to each actor, we do not need to compute the full probability \eqref{eq:prob}. Instead of using the power set of all actions, we build the power set only for the actions that are provided as weak labels for each training video clip. This means that the power set will differ for each video clip. For the example shown in Figure \ref{fig:img1}, we build the power set $\Omega$ for the actions \textit{Stand}, \textit{Listen}, \textit{Talk}, and \textit{Watch}. In this example, $\vert\Omega\vert=16$. We exclude $\varnothing$ since in the used dataset each actor is annotated with at least one action. Furthermore, we multiply $p_{\omega}$ with the confidence $d$ of the person detector. The scoring function $p_{\omega,i}$ that we use for the assignment of a subset $\omega \in \Omega \setminus \varnothing$ to a detected actor $a_i$ is therefore given by
\begin{equation}
	p_{\omega,i} = \frac{\exp\left(\sum_{c \in \omega} s_{c,i}\right) d_i}{\sum_{\omega' \in \Omega \setminus \varnothing}\exp\left(\sum_{c \in \omega'} s_{c,i}\right)} \label{eq:score}
\end{equation}
where $s_{c,i}$ is the predicted logit for action $c$ and person $a_i$. Taking the detection confidence $d_i$ of person $a_i$ into account is necessary to reduce the impact of false positives that usually have a low detection confidence.           

\subsection{Actor-Action Association}
\label{sec:asso}

While the scoring function \eqref{eq:score} indicates how likely a given subset of actions $\omega \in \Omega \setminus \varnothing$ fits to an actor $a_i$, it does not take all information that is available for each video clip into account. For instance, we know that each annotated action is performed by at least one actor. In order to exploit this additional knowledge, we find the optimal assignment of action subsets to actors based on the constraints that each actor performs at least one action and that each action $c$ occurs at least once, i.e., $c \in \bigcup_i \omega_i$. Since we build the power set only from the actions that occur in a video clip, which we denote by $L$, the power set $\Omega(L)$ varies for each training video clip. 

\begin{figure}[t]
\centering
\begin{minipage}{.31\textwidth}
\centering
   \includegraphics[width=0.9\linewidth]{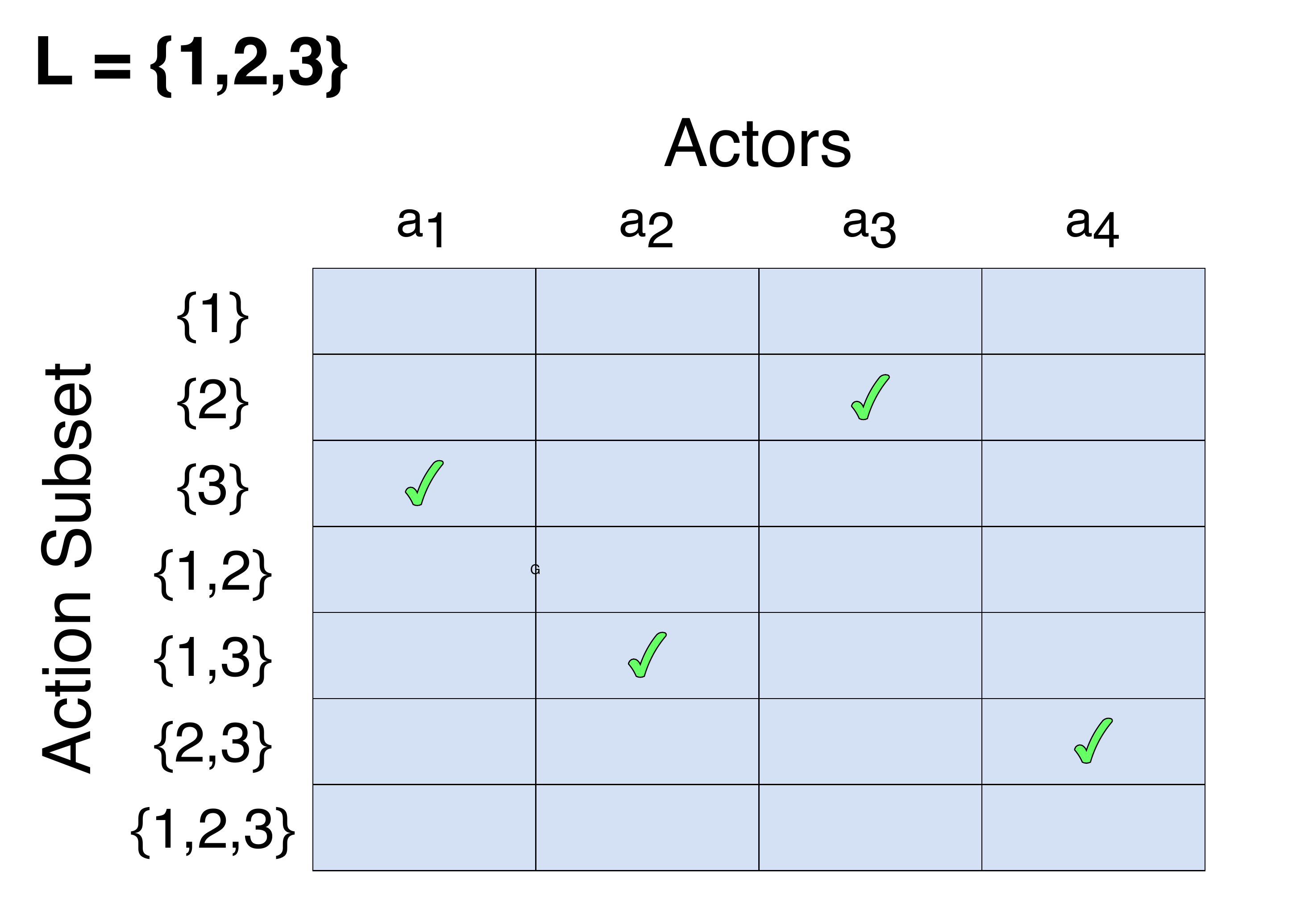}
   \text{  a) Valid assignment.}
\end{minipage}
\begin{minipage}{.31\textwidth}
\centering
   \includegraphics[width=0.9\linewidth]{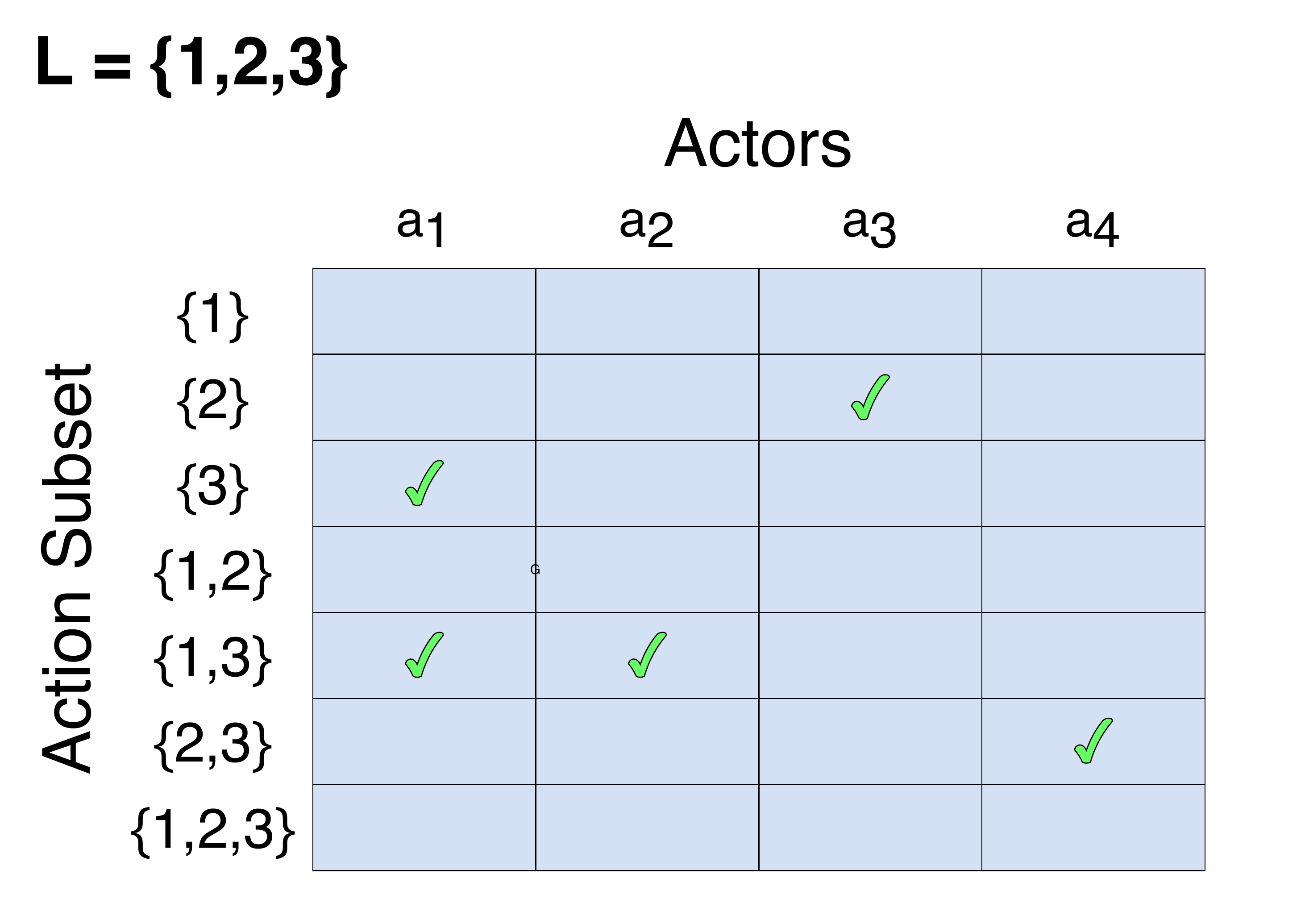}
   \text{  b) Invalid assignment.}
\end{minipage}
\begin{minipage}{.31\textwidth}
\centering
   \includegraphics[width=0.9\linewidth]{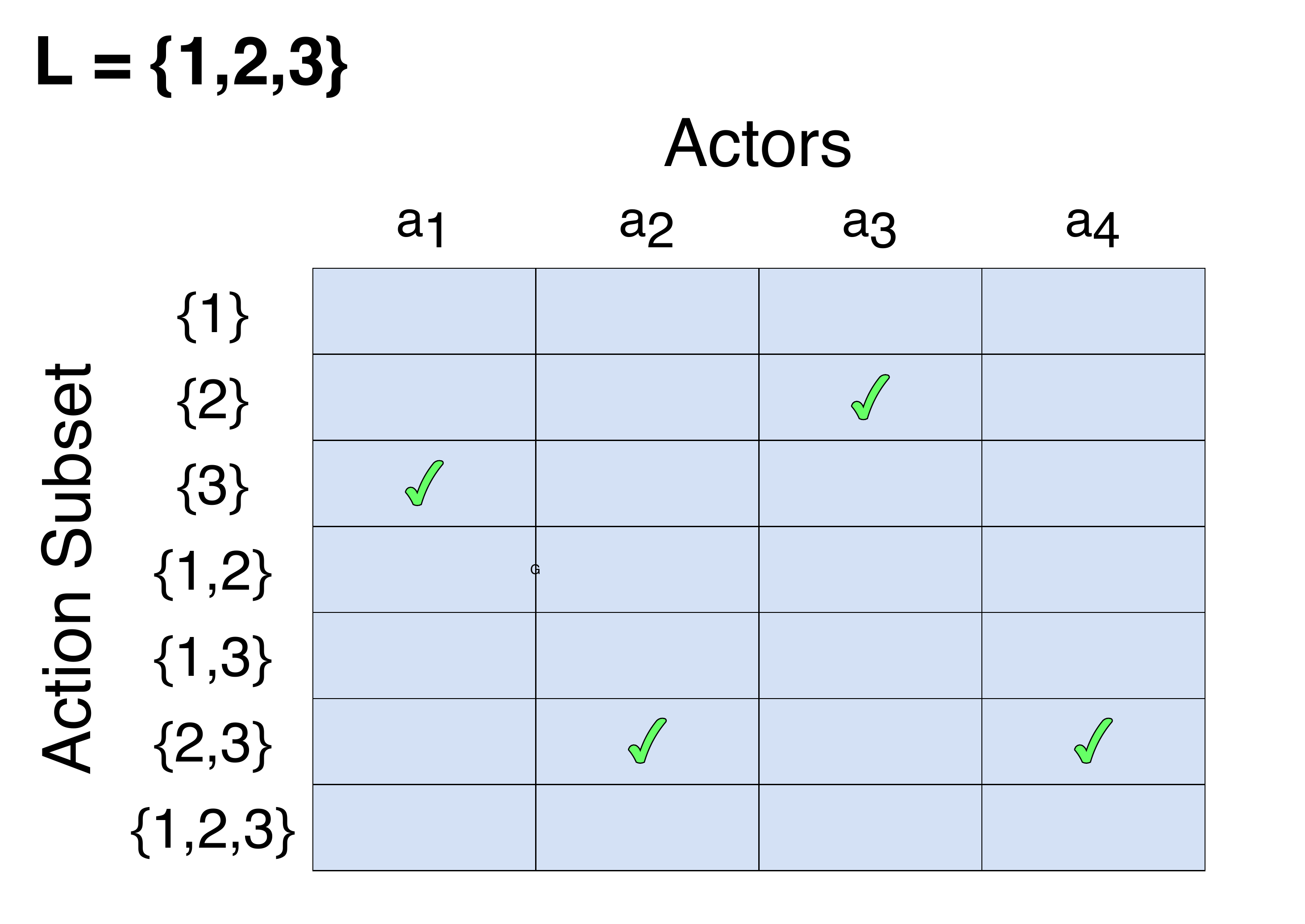}
   \text{  c) Invalid assignment.}
\end{minipage}
   \caption{For the annotated actions $L= \{1,2,3\}$ and the actors $A=\{a_1,a_2,a_3,a_4\}$, the figures demonstrate various actor-action assignments. While the assignment a) satisfies all constraints, b) violates \eqref{eq:c1} since two subsets are assigned to actor $a_1$ and c) violates \eqref{eq:c2} since the action $1$ is not part of any assigned subset.}
\label{fig:img2}
\end{figure}

The association of subsets $\omega \in \Omega(L) \setminus \varnothing$ to actors $A = \{a_1, a_2, \ldots, a_n\}$ can be formulated as a binary linear program where the binary variables $x_{\omega,i}$ are one if the subset $\omega$ is assigned to actor $a_i$ and it is zero otherwise. The optimal assignment is defined by the assignment with the highest score \eqref{eq:opti}. While the first constraint \eqref{eq:c1} enforces that exactly one subset $\omega$ is assigned to each actor $a_i$, the second constraint \eqref{eq:c2} enforces that $c \in \bigcup_{\omega : x_{\omega,i} = 1} \omega$ for all $c \in L$, where $\{\omega : x_{\omega,i} = 1\}$ is the set of all subsets that have been assigned. Note that \eqref{eq:c2} rephrases this constraint such that it can be used for optimization where the indicator function $\1{\omega}{c}$ is one if $c \in \omega$ and it is zero otherwise. The left hand side of the inequality therefore counts the number of assigned subsets that contain the action class $c$. Since this number must be larger than zero, it ensures that each action $c \in L$ is assigned to at least one actor. The complete binary linear program is thus given by:                   
\begin{alignat}{3}
 \argmax_{x_{\omega,i}}       \;  & \sum_{i=1}^{n}\sum_{\omega \in \Omega(L) \setminus \varnothing} p_{\omega,i} x_{\omega,i} &           \label{eq:opti} \\
 \text{subject to} \; & \sum_{\omega \in \Omega(L) \setminus \varnothing} x_{\omega,i}  = 1           & \forall i =1 ,..., n \label{eq:c1}\\
&                          \sum_{i=1}^{n} \sum_{\omega \in \Omega(L) \setminus \varnothing} \1{\omega}{c} x_{\omega,i}  \geq 1 & \forall c \in L   \label{eq:c2}\\
&						   x_{\omega,i}                            \in \{0,1\} &  \forall \omega \in \Omega(L) \setminus \varnothing; \; \forall i =1 ,..., n. \nonumber 
\end{alignat}
Figure~\ref{fig:img2} illustrates the constraints.

\subsection{Training}
\label{sec:train}

We train first the network using the MIML loss \eqref{eq:predopti} to obtain initial estimates of the logits $s_{c,i}$. We then assign subsets of actions to the detected persons using the scoring function \eqref{eq:score}. Finally, we train our network using the loss 
\begin{equation}
\label{eq:overallloss}
	\mathcal{L} = \mathcal{L}_{MIML} + \alpha \sum_{i=1}^{n_t} \mathcal{L}\left(\hat{Y}_{\omega^t_i}, f(a^t_{i})\right) 
\end{equation}
where $\omega^t_i$ denotes the action subset that has been assigned to actor $a^t_i$ in frame $t$ and $\hat{Y}_{\omega^t_i}$ is a vector with $\hat{Y}_{\omega^t_i}(c) = 1$ if $c \in \omega^t_i$ and $\hat{Y}_{\omega^t_i}(c) = 0$ otherwise. $\mathcal{L}$ is the binary cross entropy. Since $\mathcal{L}_{MIML}$ is computed once per frame but $\mathcal{L}(\hat{Y}_{\omega^t_i}, f(a^t_{i}))$  is computed for each detected person, we use $\alpha=0.3$ to compensate for this difference.  

\section{Experiments}
\subsection{Dataset and Implementation Details}

We use the AVA 2.2 dataset \cite{gu2018ava} for evaluation. The dataset contains 235 videos for training, 64 videos for validation, and 131 videos for testing. The dataset contains 60 action classes. The persons perform often multiple actions at the same time and the videos contain multiple persons. For each annotated person a bounding box is provided. An example is given in Figure~\ref{fig:img1}. Only one frame per second is annotated. The accuracy is measured by mean average precision (mAP) over all actions with an IoU threshold for bounding boxes of 0.5 as described in \cite{gu2018ava}. In the weakly supervised setting, we use only the present actions for training, but not the bounding boxes.

To detect persons, we use Faster RCNN \cite{ren2015faster} with ResNext-101 \cite{xie2017resnext} as backbone. The detector was pre-trained on ImageNet and fine-tuned on the COCO dataset. In our experiments, we report results for two 3D CNNs, namely I3D \cite{i3d} and Slowfast \cite{feichtenhofer2019slowfast}. I3D is pre-trained on Kinetics-400. For Slowfast, we use the ResNet-101 + NL ($8\times8$) version that is pre-trained on Kinetics 600. The temporal scope was set to 64 frames with a stride of 2. For HGRNN we use a temporal window of $11$ frames. For training, we use the SGD optimizer until the  validation error saturated. The learning rate with linear warmup was set to 0.04 and 0.025 for I3D and Slowfast, respectively. The batch size was set to 16. We used cropping as data augmentation where we crop images of size $224 \times 224$ pixels from the frames that have $256 \times 256$ image resolution.\footnote{Code: https://github.com/sovan-biswas/MultiLabelActorActionAssignment} 

\subsection{Experimental Results} 

\begin{table}[t]
\caption{Comparison of MIML with proposed method. The proposed approach outperforms MIML in case of I3D and Slowfast.}
\centering
	    \begin{tabular}{c|c|c}
            Method & 3D CNN & Val-mAP \\
            \hline
            MIML & I3D & 14.1 \\
           	MIML + HGRNN  & I3D &  15.2 \\
            Proposed Approach & I3D &  {17.3} \\
            \hline
            MIML & Slowfast & 21.8  \\
           	MIML + HGRNN & Slowfast &  23.1 \\
            Proposed Approach & Slowfast &  {25.1} \\
            \hline
	    \end{tabular}
		\label{tab:base}
\end{table}

\subsubsection*{Comparison of MIML with proposed method.} 
\begin{figure}[t]
\centering
   \includegraphics[trim=0cm 0cm 0cm 0cm, clip,width=0.7\linewidth]{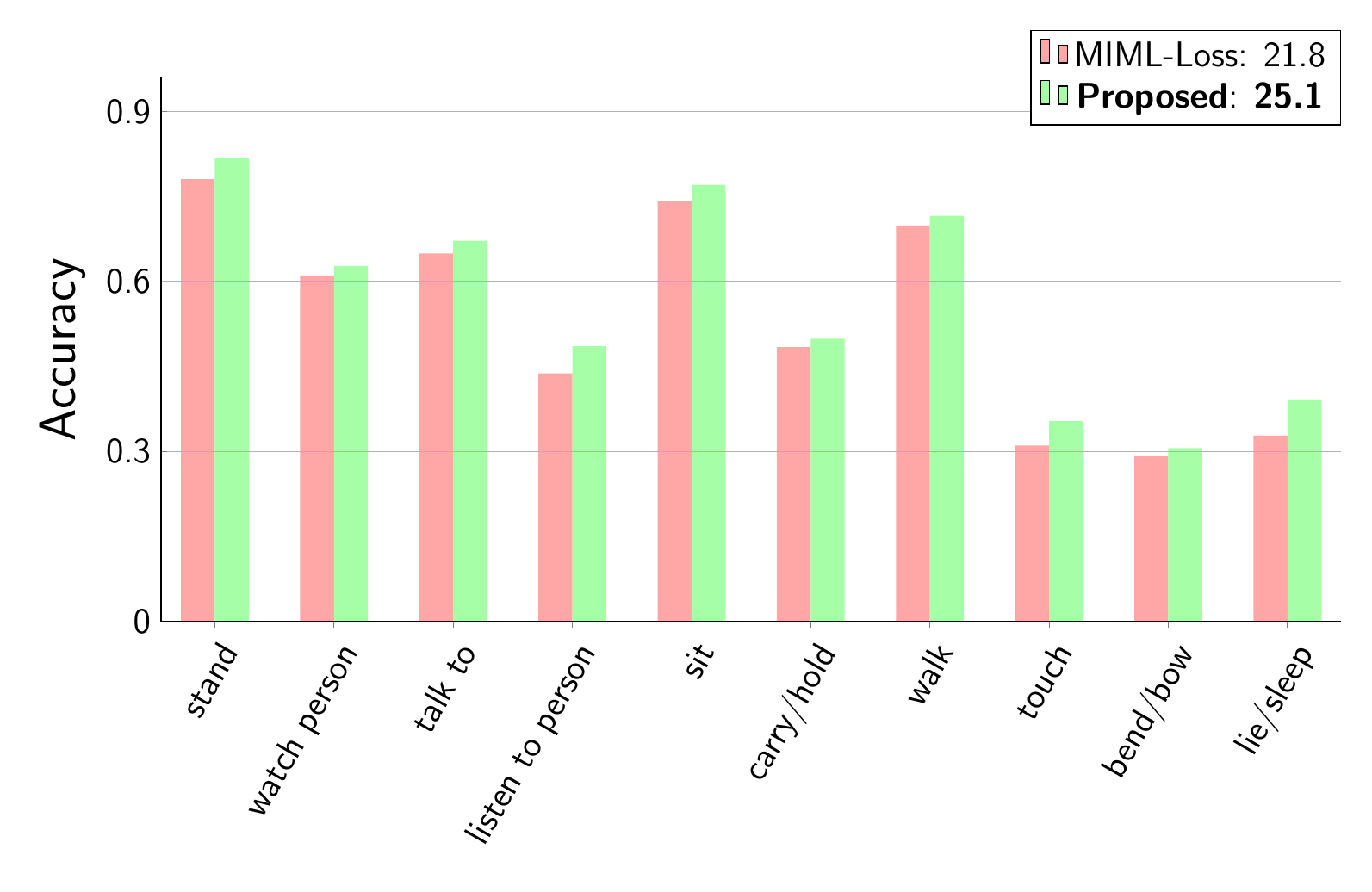}
   \caption{Comparison of MIML with proposed method. The plot shows the per class mAP for the 10 most frequently occurring classes in the training set. The actions are sorted by the number of occurrences in an decreasing order from left to right. A plot with all 60 action classes is part of the supplementary material.}
\label{fig:bar_plot}
\end{figure}

\begin{figure}[th]
\begin{minipage}{.325\textwidth}
\centering {\normalsize {Ground-Truth}}
\end{minipage}
\begin{minipage}{.325\textwidth}
\centering {\normalsize {MIML}}
\end{minipage}
\begin{minipage}{.325\textwidth}
\centering {\normalsize {Proposed}}
\end{minipage}
\\
\hrule

\begin{minipage}{.325\textwidth}
\centering
  \includegraphics[trim=1.8cm 1.5cm 2cm 0cm, clip,width=0.67\linewidth]{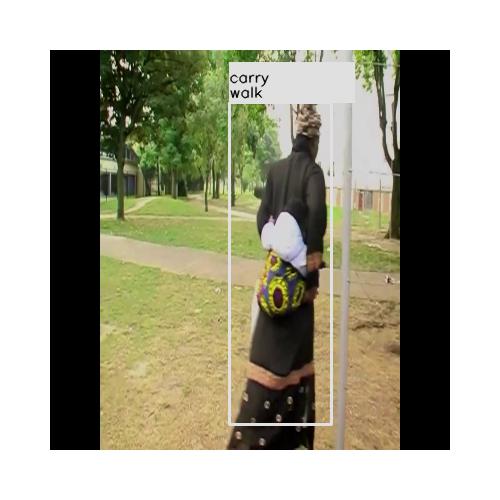}
\end{minipage}
\begin{minipage}{.325\textwidth}
\centering
  \includegraphics[trim=1.8cm 1.5cm 2cm 0cm, clip,width=0.67\linewidth]{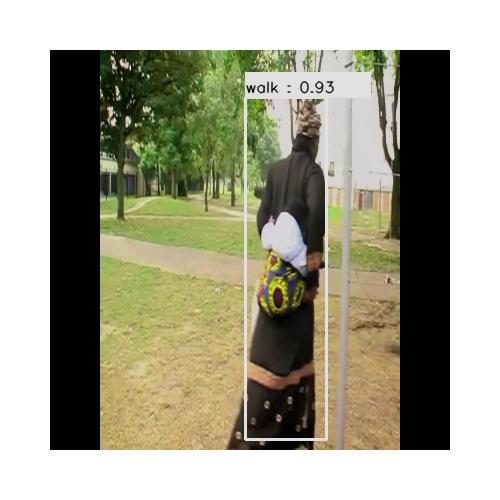}
\end{minipage}
\begin{minipage}{.325\textwidth}
\centering
  \includegraphics[trim=1.8cm 1.5cm 2cm 0cm, clip,width=0.67\linewidth]{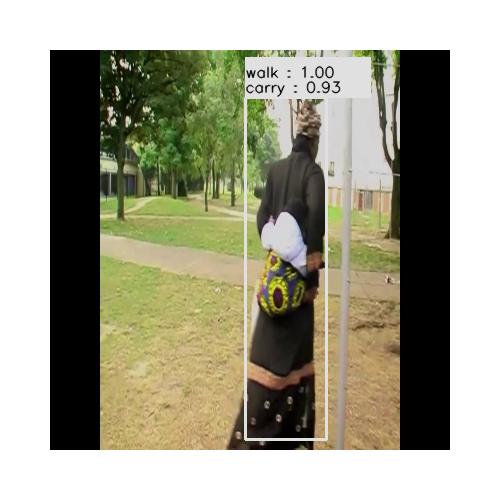}
\end{minipage}

\begin{minipage}{.325\textwidth}
\centering
  \includegraphics[trim=1.8cm 1.5cm 2cm 0cm, clip,width=0.67\linewidth]{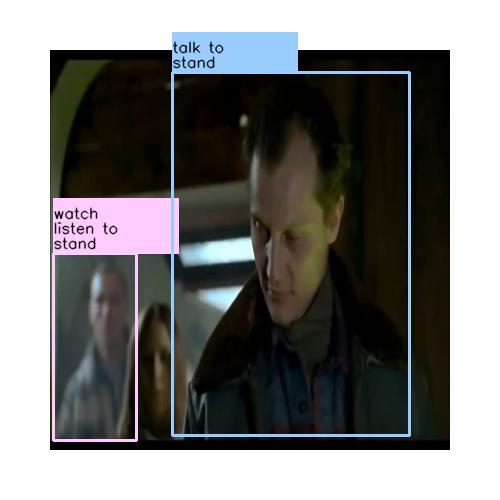}
\end{minipage}
\begin{minipage}{.325\textwidth}
\centering
  \includegraphics[trim=1.8cm 1.5cm 2cm 0cm, clip,width=0.67\linewidth]{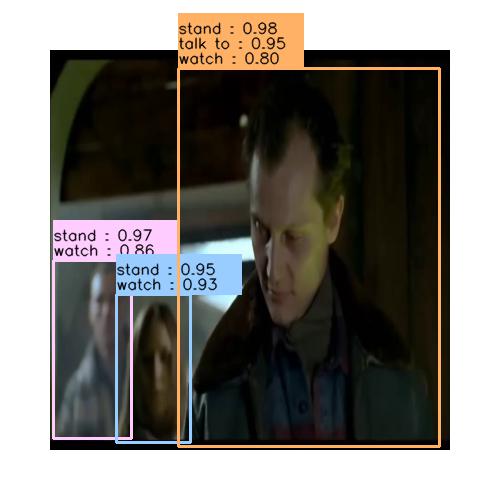}
\end{minipage}
\begin{minipage}{.325\textwidth}
\centering
  \includegraphics[trim=1.8cm 1.5cm 2cm 0cm, clip,width=0.67\linewidth]{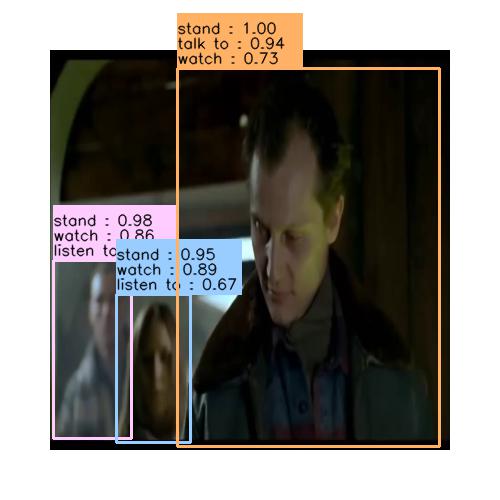}
\end{minipage}

\begin{minipage}{.325\textwidth}
\centering
  \includegraphics[trim=1.8cm 1.5cm 2cm 0cm, clip,width=0.67\linewidth]{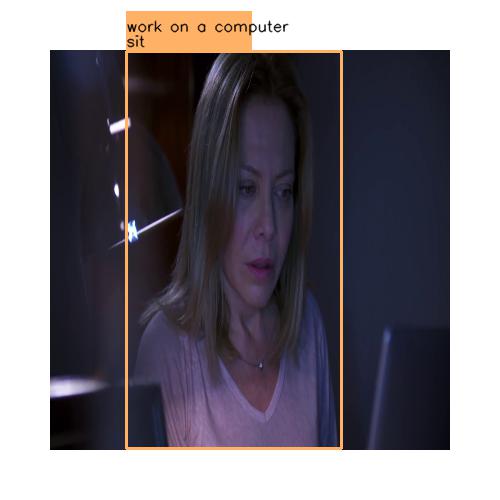}
\end{minipage}
\begin{minipage}{.325\textwidth}
\centering
  \includegraphics[trim=1.8cm 1.5cm 2cm 0cm, clip,width=0.67\linewidth]{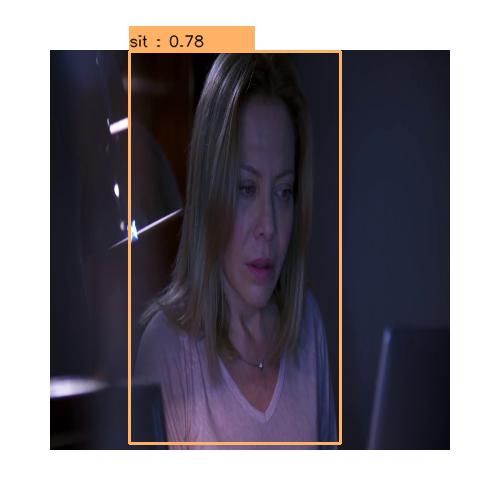}
\end{minipage}
\begin{minipage}{.325\textwidth}
\centering
  \includegraphics[trim=1.8cm 1.5cm 2cm 0cm, clip,width=0.67\linewidth]{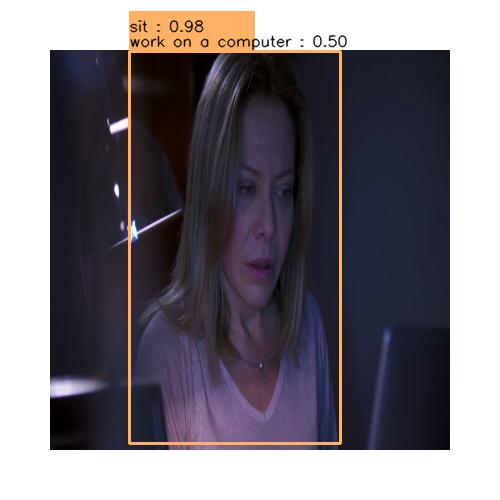}
\end{minipage}

\begin{minipage}{.325\textwidth}
\centering
  \includegraphics[trim=1.8cm 1.5cm 2cm 0cm, clip,width=0.67\linewidth]{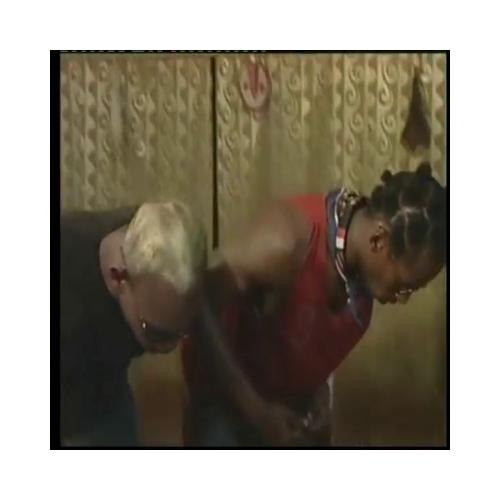}
\end{minipage}
\begin{minipage}{.325\textwidth}
\centering
  \includegraphics[trim=1.8cm 1.5cm 2cm 0cm, clip,width=0.67\linewidth]{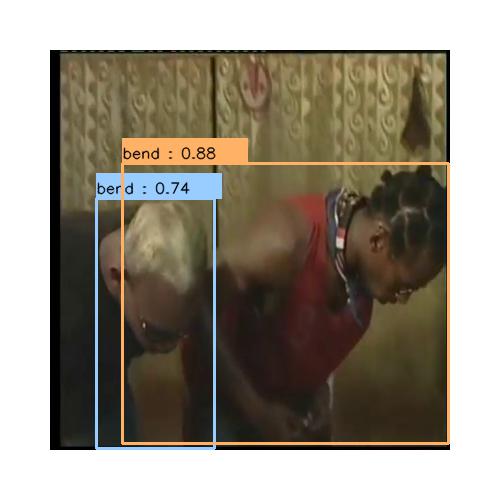}
\end{minipage}
\begin{minipage}{.325\textwidth}
\centering
  \includegraphics[trim=1.8cm 1.5cm 2cm 0cm, clip,width=0.67\linewidth]{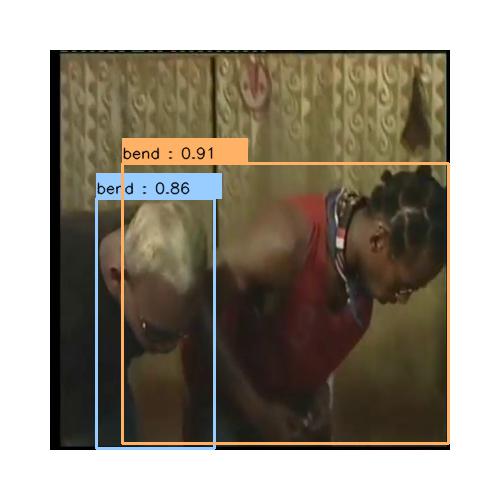}
\end{minipage}

\begin{minipage}{.325\textwidth}
\centering
  \includegraphics[trim=1.8cm 1.5cm 2cm 0cm, clip,width=0.67\linewidth]{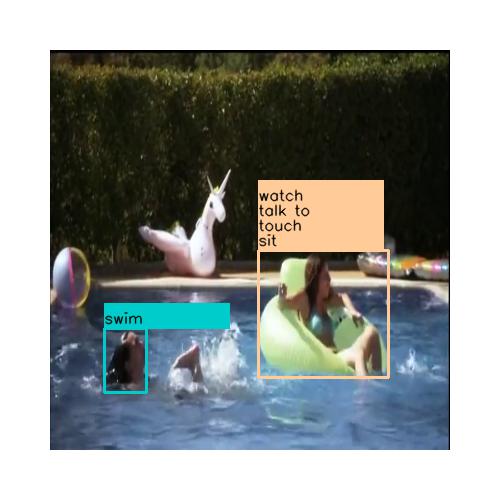}
\end{minipage}
\begin{minipage}{.325\textwidth}
\centering
  \includegraphics[trim=1.8cm 1.5cm 2cm 0cm, clip,width=0.67\linewidth]{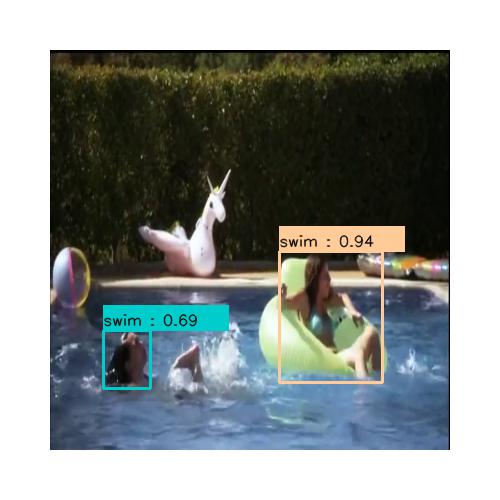}
\end{minipage}
\begin{minipage}{.325\textwidth}
\centering
  \includegraphics[trim=1.8cm 1.5cm 2cm 0cm, clip,width=0.67\linewidth]{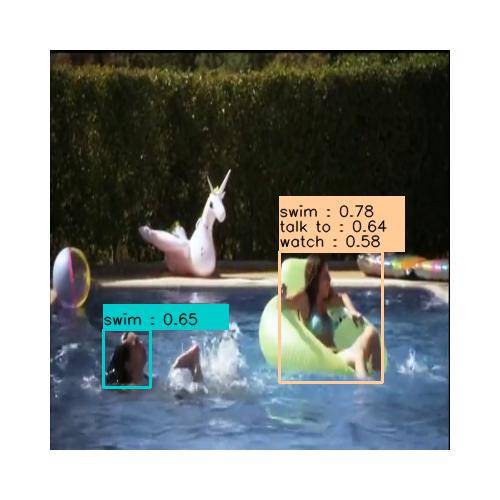}
\end{minipage}

\caption{Qualitative results. The left column shows the ground-truth annotations. The middle column shows the results of the MIML baseline. The right column shows the results of the proposed method. The colors distinguish only different persons, but they are otherwise irrelevant. The predicted action classes with confidence scores are on top of the estimated bounding boxes. The proposed approach recognizes more action classes per bounding box correctly compared to MIML. Both methods also detect genuine actions that are not annotated in the dataset as seen from the missing persons in the second and fourth row. The bias of the proposed method towards the background is visible in last row, where the ``swim'' action is associated to both persons. Best viewed using the zoom function of the PDF viewer.}
\label{fig:Actor-ActionRefinement}
\end{figure}

Table~\ref{tab:base} shows the comparison of the proposed approach with the multi-instance and multi-label (MIML) baseline on the validation set. When I3D is used as 3D CNN, the proposed approach improves the MIML baseline by $+3.2\%$. When Slowfast is used, the accuracy of all methods is higher but the improvement of the proposed approach over the MIML approach remains nearly the same with $+3.3\%$. We also report the result when HGRNN is trained only with the MIML loss. In this case, the actor-action association is not used and we denote this setting by MIML+HGRNN. While HGRNN improves the results since it models the spatio-temporal relations between persons better than a 3D CNN alone, the proposed actor-action assignment improves the mAP compared to MIML+HGRNN by $+2.1\%$ and $+2.0\%$ for I3D and Slowfast, respectively. Figure~\ref{fig:bar_plot} shows the improvement of the proposed approach over the MIML baseline for the 10 action classes that occur most frequently in the training set. A few qualitative results are show in Figure~\ref{fig:Actor-ActionRefinement}.

\begin{table}[t]
	    \caption{Results of various actor-action assignment approaches using HGRNN over different 3D CNNs. The Frequent-5 column and the Least-10 column show the average mAP over the 5 most frequently and 10 least occurring classes in the training set.}
\centering
	  \begin{tabular}{c|c|c|c|c}
            Actor-action association & Backbone & Val-mAP & Frequent-5 & Least-10 \\
            \hline
            MIML+HGRNN  & I3D & 15.2 & 51.5 & 2.0 \\
           	Proposed Approach w/o LP & I3D & 16.4 & 52.8 & 2.1 \\
            Proposed Approach  & I3D & \textbf{17.3} & \textbf{53.7} & \textbf{3.4}\\
            \hline
            MIML+HGRNN  & Slowfast & 23.1 & 65.7 & 7.3 \\
           	Proposed Approach  w/o LP & Slowfast & 22.9 & 65.9 & 6.8 \\
            Proposed Approach  & Slowfast & \textbf{25.1} & \textbf{67.5} & \textbf{7.6}\\
            \hline
	    \end{tabular}
\label{tab:actor-action}
\end{table}

\subsubsection*{Impact of actor-action association.} In Table \ref{tab:base}, we have observed that the actor-action association improves the accuracy. In Table~\ref{tab:actor-action}, we analyze the impact of the actor-action association more in detail. We use HGRNN using both I3D and Slowfast as 3D CNN backbone. In case of MIML+HGRNN, the actor-action association is not used. We also report the result if we perform the association directly by the confidences without solving a binary linear program. We denote this setting by Proposed Approach w/o LP. In this case, we associate an action to an actor if the class probability is greater than 0.5. For I3D, the association without LP improves the results mainly for the most frequent classes with almost no improvement on least frequent classes. For Slowfast, the performance even decreases in comparison to MIML+HGRNN without LP. Instead, solving the linear program results in better associations for both I3D and Slowfast. 

\begin{table}[t]
  \caption{Performance with ground-truth bounding boxes for evaluation. The results show the improvement in mAP on the validation set when ground-truth bounding boxes (GT bb) instead of detected bounding boxes (Detected bb) are used for evaluation. Furthermore, the results are reported when the model is trained with full supervision.}
\centering
	    \begin{tabular}{c|c|c|c}
            Method & 3D CNN & Detected bb & GT bb\\
            \hline
           	MIML & I3D & 14.1 & 21.2\\
            Proposed Approach & I3D & {17.3} & {24.3} \\
            Full Supervision & I3D & {20.7} & {25.4} \\
            \hline \hline
            MIML & Slowfast & 21.8 & 30.6\\
            Proposed Approach & Slowfast &  {25.1} & {32.3} \\
            Full Supervision & Slowfast & {30.1} & {35.7} \\
            \hline
	    \end{tabular}
		\label{tab:onGT}
\end{table}

\subsubsection*{Impact of the object detector.}
We use the Faster RCNN with ResNext \cite{xie2017resnext} person detector which achieves $90.10\%$ mAP for person detection on the AVA training set and $90.45\%$ on the AVA validation set. Irrespective of the high detection performance, we analyze how much the accuracy improves if the detected bounding boxes are replaced with the ground-truth bounding boxes during evaluation. Note that the ground-truth bounding boxes are not used for training, but only for evaluation. The results are shown in Table~\ref{tab:onGT}. We observe that the performance improves by $+7.0\%$ and $+7.2\%$ mAP on the validation set for I3D and Slowfast, respectively. We also report the results if the approach is trained using full supervision. In this case, the network is trained on the ground-truth bounding boxes and the ground-truth action labels per bounding box. Compared to the fully supervised approach, our weakly supervised approach achieves around $83\%$ of the mAP for both 3D CNNs ($17.3\%$ vs.\ $20.7\%$ for I3D and $25.1\%$ vs.\ $30.1\%$ for Slowfast) if detected bounding boxes are used for evaluation. The gap gets even smaller when ground-truth bounding boxes are used for evaluation. In this case, the relative performance is $95.7\%$ for I3D and $90.5\%$ for Slowfast. This demonstrates that the proposed approach learns the actions very well despite of the weak supervision. 

\begin{table}[t]
\centering
\caption{Comparison to fully supervised approaches. We also report the result of our approach if it is trained with full supervision. Note that we do not use multi-scale and horizontal flipping augmentation as in Slowfast++.}
	    \begin{tabular}{c|c|c}
            
            \multicolumn{3}{c}{Weakly Supervised Approaches}\\
            \hline
            Methods & Val-mAP & Test-mAP\\
            \hline
            MIML&21.8&-\\
            Proposed Approach & \textbf{25.1} & \textbf{23.5}\\
            \hline \hline
            \multicolumn{3}{c}{Fully Supervised Approaches}\\
            \hline
            Methods & Val-mAP & Test-mAP\\
            \hline
            ARCN \cite{sun2018actor} & 17.4 & - \\
            RAF \cite{Sun_2019_CVPR} & 20.4 & - \\
            HGRNN \cite{biswas2019} & 20.9 & - \\
            ATN \cite{girdhar2019video} & 25.0 & 24.9 \\
            LFB \cite{wu2019long} & 27.7 & {27.2}\\
            Slowfast \cite{feichtenhofer2019slowfast} & {29.0}  & - \\
            Slowfast++ \cite{feichtenhofer2019slowfast} & {30.7}  & {34.3} \\
            Proposed Approach & 30.1 & - \\
            \hline
	    \end{tabular}
		\label{tab:weakVsSuper}
\end{table}

\subsubsection*{Comparison to fully supervised methods.} Since this is the first approach that addresses weakly supervised learning for multi-label and multi-person action detection, we cannot compare to other weakly supervised approaches. However, we compare our approach with the state-of-the-art for fully supervised action detection in Table~\ref{tab:weakVsSuper}. Our approach is competitive to fully supervised approaches \cite{sun2018actor,Sun_2019_CVPR,biswas2019,girdhar2019video}. When we train our approach with full supervision, we improve over SlowFast \cite{feichtenhofer2019slowfast} by $+1.1\%$ mAP on the validation set. While the Slowfast++ network performs slightly better, it uses additional data augmentation and a different network configuration. We expect that these changes would improve our approach as well.          

\section{Conclusion}
In this paper, we introduced the challenging task of weakly supervised multi-label spatio-temporal action detection with multiple actors. We first introduced a baseline based on multi-instance and multi-label learning. We furthermore presented a novel approach where the multi-label problem is represented by the power set of the action classes. In this context, we assign an element of the power set to each detected person using linear programming. We evaluated our approach on the challenging AVA dataset where the proposed method outperforms the MIML approach. Despite of the weak supervision, the proposed approach is competitive to fully supervised approaches.     
 
\subsection*{Acknowledgment}
The work has been financially supported by the ERC Starting Grant ARCA (677650).
\bibliographystyle{splncs}
\bibliography{egbib}
\end{document}